\documentclass[10pt,twocolumn,letterpaper]{article}

\usepackage{iccv}
\usepackage{times}
\usepackage{epsfig}
\usepackage{graphicx}
\usepackage{amsmath}
\usepackage{amssymb}
\usepackage{enumitem}

\usepackage[pagebackref=true,breaklinks=true,letterpaper=true,colorlinks,bookmarks=false]{hyperref}

\iccvfinalcopy 

\def\iccvPaperID{11189} 

\ificcvfinal\fi

\begin{document}



\title{TransForensics: Image Forgery Localization with Dense Self-Attention}


\author{Jing Hao\footnotemark[1] \qquad Zhixin Zhang\footnotemark[1] \qquad Shicai Yang \qquad Di Xie \qquad Shiliang Pu\footnotemark[2]\\
Hikvision Research Institute\\
{\tt\small \{haojing, yangshicai, xiedi, pushiliang.hri\}@hikvision.com}
}

\maketitle

\renewcommand{\thefootnote}{\fnsymbol{footnote}}
\footnotetext[1]{Equal contribution.}
\footnotetext[2]{Corresponding author.}

\ificcvfinal\thispagestyle{empty}\fi

\begin{abstract}
Nowadays advanced image editing tools and technical skills produce tampered images more realistically, which can easily evade image forensic systems and make authenticity verification of images more difficult. To tackle this challenging problem, we introduce TransForensics, a novel image forgery localization method inspired by Transformers. The two major components in our framework are dense self-attention encoders and dense correction modules. The former is to model global context and all pairwise interactions between local patches at different scales, while the latter is used for improving the transparency of the hidden layers and correcting the outputs from different branches. Compared to previous traditional and deep learning methods, TransForensics not only can capture discriminative representations and obtain high-quality mask predictions but is also not limited by tampering types and patch sequence orders. By conducting experiments on main benchmarks, we show that TransForensics outperforms the state-of-the-art methods by a large margin.

\end{abstract}

\section{Introduction}

Image is an important medium for information transmission. Recently, tampered images generated by image editing techniques are commonly confused to be real ones, and are increasingly used in fake news creation, academic fraud, and criminal offenses. When tampering occurs in a digital image, we usually expect that the tampered regions can be found through image forensic analyses. However, capturing discriminative features of tampered regions with multiple forgery types (\eg splicing, copy-move, removal) is still a challenge and often requires exploiting the characteristics of different tampering artifacts.

\begin{figure}[t]
  \centering
  \includegraphics[width=8.2cm]{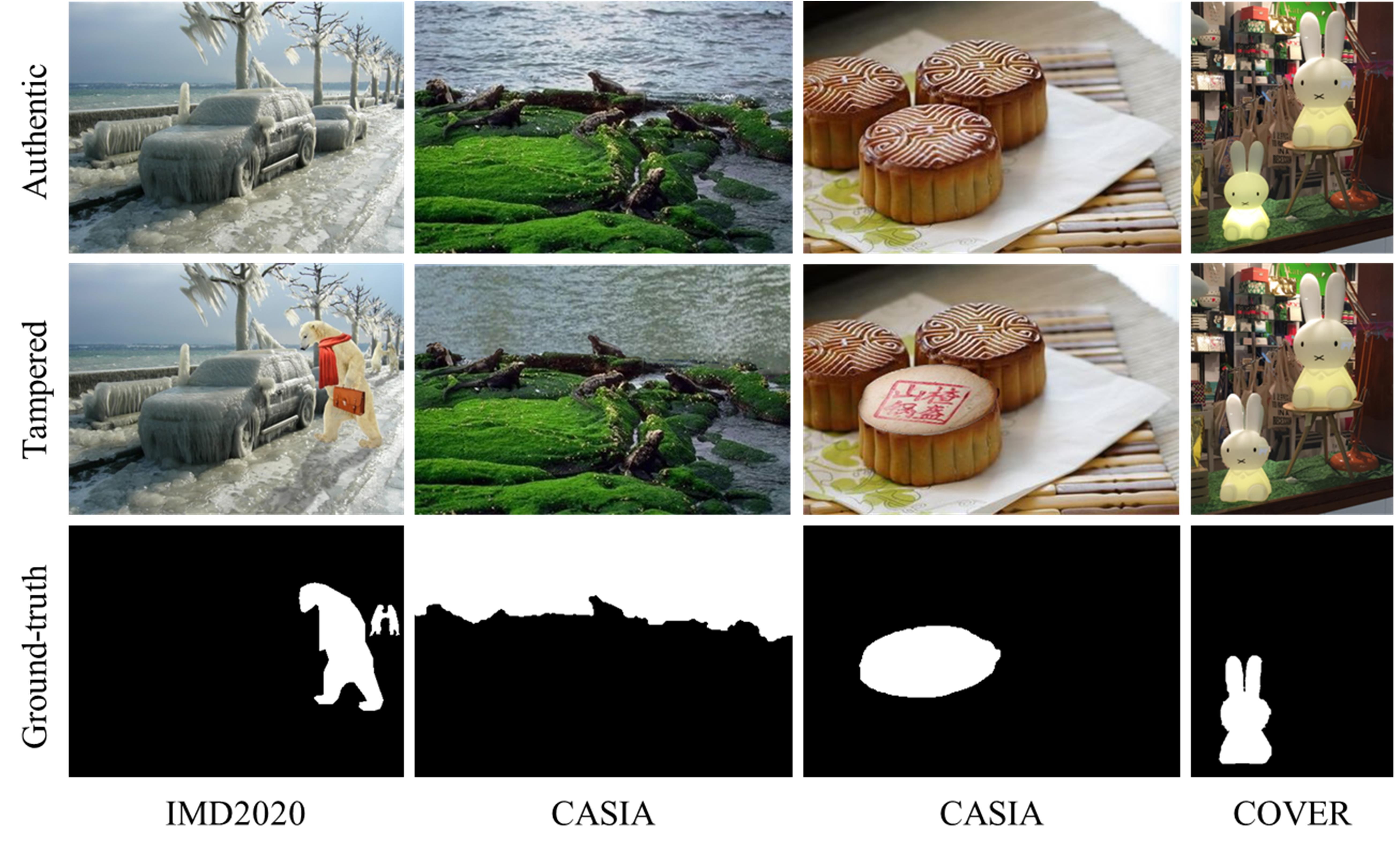}
  \caption{Examples from three common image forgery datasets. Four authentic images (top) with their corresponding tampered images (medium) and ground-truth masks (bottom).}
  \label{fig:example}
\end{figure}

Unlike semantic object segmentation methods \cite{7478072, DBLP:journals/corr/abs-1807-10165} that do predictions of all meaningful object regions, image forensics makes predictions of tampering locations only. The former focuses on analyzing the content of different regions to understand visual concepts, while the latter needs to generalize some other artifacts (\eg inconsistent local noise variances) created by different manipulation techniques. As shown in Fig.\ref{fig:example}, the well-manipulated images are usually realistic, where the content of fake and genuine regions is likely to be similar. If we directly use semantic segmentation network for image forensics, the network would localize both original and manipulated regions, while the original ones are the wrong predictions for image forensics. The work of Bappy \etal \cite{bappy2017exploiting} also have demonstrated that semantic segmentation approaches do not perform well for image manipulations.

The key to image forensics is characterizing different tempering artifacts that are often hidden in tiny details of the images. Previous methods mainly employ traditional hand-crafted features, such as error level analysis (ELA) \cite{krawetz2007picture}, discrete cosine transform (DCT) \cite{barni2017aligned}, and steganalysis rich model (SRM) \cite{fridrich2012rich}, to learn local inconsistencies from invisible traces, but they usually apply only to a specific manipulation type. In fact, the boundary formation of tampered (smoother) and authentic regions (sharper) within an image is different \cite{bappy2017exploiting, bappy2019hybrid}. With the success of deep learning approaches, recent works focus on checking feature consistency or learning boundary discrepancy via convolutional neural networks (CNNs) \cite{cozzolino2018noiseprint, mazumdar2018universal, verde2020focal, zhou2018learning} or recurrent neural networks (RNNs) \cite{bappy2017exploiting, bappy2019hybrid}, which allow to capture tampering features and perform better than traditional methods.

However, the major shortcoming of deep learning methods is that they heavily depend on hand-designed patch sequence orders and manipulation types. Specifically, RNNs based methods split an image into a series of patches and use a long-short term memory (LSTM) network to learn the correlations between them. These networks can receive the sequential inputs, but cannot retain the spatial location information. In contrast, the methods combining hand-crafted features with deep features \cite{bammey2020adaptive, wu2019mantra, zhou2017two, zhou2018learning} achieve the state-of-the-art (SOTA) performance, but they usually assume that tampering type is known beforehand. Taking these facts into consideration, here we show how a spatial attention network can be used within an image forgery localization framework to model all pairwise interactions between patches (including rich statistical features) of an image, yet maintain the global structure and alleviate ordering techniques and manipulation types limitation. \medskip

\noindent{\bf Framework overview}\quad In this paper, the goal of our system is to predict binary masks for image forgery localization. Firstly, we use a fully convolutional network (FCN) as backbone for feature extraction. Then, self-attention encoders are used to model rich interactions between points in feature maps at different scales. For improving the performance, dense correction modules are used in our network, which helps to learn more discriminative representations from the early layers and performs results correction. \medskip

\noindent{\bf Main contributions}\quad In this work, the main contributions are as follows. First, we propose a novel image forgery localization method, called TransForensics. To the best of our knowledge, it is the first attempt in image forensics to model all pairwise relations, yet maintain the spatial structure between patches with the self-attention mechanism. Second, we introduce a dense correction architecture, which adds the direct supervision for the hidden layers, and corrects the outputs from different branches by multiplication. Experiments show that our method outperforms the SOTA methods by a large margin. \medskip

\noindent{\bf Structure of the paper}\quad The paper is organized as follows. We first review related work in Section 2. Then, Section 3 introduces the proposed dense attention network for image forgery localization in detail. Section 4 shows the experimental datasets, details, results and analysis. Finally, Section 5 gives the conclusion of this paper.

\section{Related work}

Prior work on which our work build contains several domains: image forensics, self-attention mechanism, and deep supervision. Image forensics focuses on detecting tampering artifacts, and research on this domain contains various traditional and deep learning approaches for image forgery classification, detection, and localization. The self-attention mechanism is the core component of Transformers, which are widely used in nature language processing (NLP) and computer vision (CV). Deep supervision attempts to enforce direct supervision for the hidden layers, where the learned features are sensible and discriminative. In this section, we briefly review prior work.\medskip

\begin{figure*}[t]
  \centering
  \includegraphics[width=15cm]{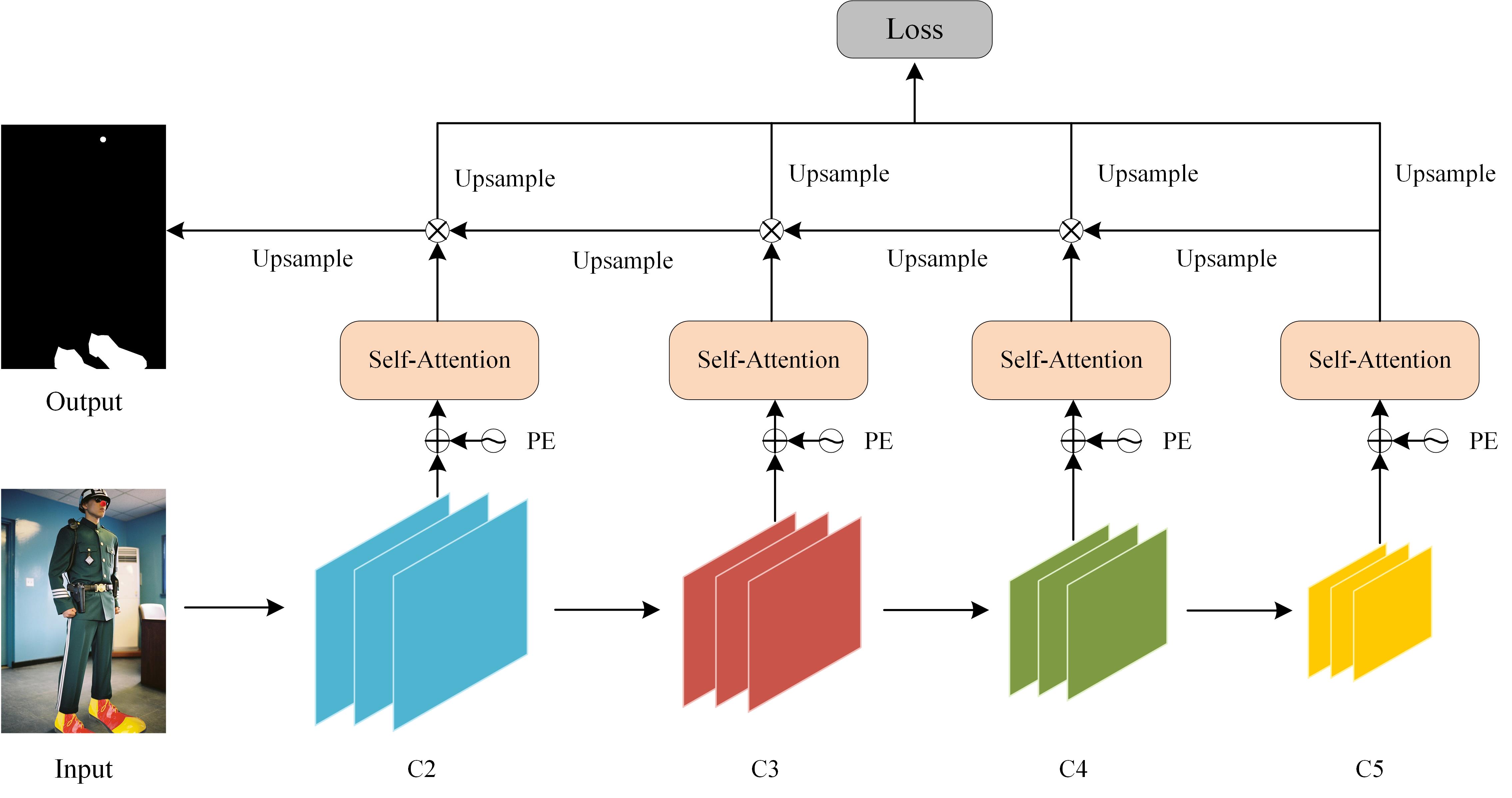}
  \caption{Architecture of image forgery localization network. The whole image is the input signal. First, a FCN backbone is applied to extract discriminative features. Then, the features from four blocks ($C_2$, $C_3$, $C_4$ and $C_5$) combining with positional encodings are input into the self-attention encoders separately, which captures rich interactions between `patches' in the input image. Finally, a feature fusion strategy by multiplication corrects the mask predictions. In this work, we do not split the whole image into a series of patches, and the points in feature maps is equivalent to the invisible patches in the input image (see Fig. \ref{fig:viewpatch}).}
  \label{fig:network}
\end{figure*}

\noindent{\bf Image forensics}\quad The development of image editing techniques makes tampered images widely available and more realistic. The most common tampering techniques are splicing, copy-move, and removal. Splicing means copying regions from an image to another image. Copy-move copies and pastes regions within the same image. Removal means removing regions from the current image. Image forensics aims at detecting these tampering artifacts, and involves binary (real or fake) classification and tampered regions detection or localization tasks. At first, many studies in this filed are traditional methods based on hand-crafted features, such as local noise analysis \cite{cozzolino2014image, fridrich2012rich, lyu2014exposing, pan2012exposing}, CFA artifacts \cite{ferrara2012image}, illumination variance analysis \cite{de2013exposing, riess2010scene}, and double JPEG compression \cite{barni2017aligned, bianchi2011improved, krawetz2007picture}. With the revolutionary advance of deep learning, some methods try to bring deep neural networks into this realm (\eg RNNs \cite{bappy2017exploiting, bappy2019hybrid}, CNNs \cite{cozzolino2018noiseprint, hu2020span, li2020face, verde2020focal} and GANs \cite{islam2020doa}). There are also many papers that combine hand-crafted features and deep features for image forensics \cite{amerini2017localization, bammey2020adaptive, wang2016double, wu2019mantra, zhou2017two, zhou2018learning}.

Local noise variances estimation is used for image splicing detection \cite{pan2012exposing}. This is because different regions within an authentic image containing intrinsic noise have similar noise variances, and image splicing can be exposed with inconsistent local noise variances. Similarly, SRM \cite{fridrich2012rich, zhou2017two, zhou2018learning} uses local noise residuals to capture the inconsistency between tampered and authentic regions. For example, Fridrich \etal \cite{fridrich2012rich} propose steganalyzers to construct rich models of the noise component to capture numerous quantitative relationships between pixels in an image, and Zhou \etal \cite{zhou2017two} combine SRM features with RGB features by a two-stream Faster R-CNN to perform manipulation detection. Furthermore, Bammey \etal \cite{bammey2020adaptive} design a CNN structure based on demosaicing algorithms to point out local mosaic inconsistencies. Amerini \etal \cite{amerini2017localization} combines a spatial domain CNN with a frequency domain CNN for splicing forgery detection, which is inspired by the fact that the artifacts of single and double JPEG compression are different. Bappy \etal \cite{bappy2017exploiting} propose to exploit the interdependency between patches, which is efficient for various types of manipulation detection, and then they present a hybrid CNN-LSTM network \cite{bappy2019hybrid} utilizing resampling features to improve the detection performance. The studies combining hand-crafted features with deep features are: CNN and CFA \cite{bammey2020adaptive}, CNN and steganalysis \cite{wu2019mantra, zhou2017two, zhou2018learning}, CNN and double JPEG compression \cite{amerini2017localization, wang2016double}.

The commonality among all traditional algorithms discussed above is that they usually apply only to a certain manipulation type, and the disadvantage of deep learning methods is that the performance heavily depends on patch ordering techniques. Motivated by these works, here we present a novel spatial attention network with the self-attention mechanism for modeling rich interactions between patches at different scales. \medskip

\noindent{\bf Self-attention mechanism}\quad The self-attention mechanism is the core component of Transformers \cite{vaswani2017attention}, which has been successfully used in CV \cite{khan2021transformers}. Here, self-attention can capture `long-term' dependencies between set elements (\eg pixels, image patches or video frames) to aggregate global information of the input signal. A recent framework, called DETR \cite{nicolas2020transformers}, views object detection as a direct set prediction problem and uses Transformers with parallel decoding to produce unique predictions. Semantic segmentation is a dense prediction task, where Transformers can be used to model relations between pixels. For example, Ye \etal \cite{ye2019crossmodal} propose a cross-modal self-attention to learn long-range dependencies between linguistic and visual features, and Zheng \etal \cite{zheng2020rethinking} deploy a pure Transformer to encode an image as a sequence of patches. However, the usual solution for image forensics to learn the correlations between patches is using LSTM cells, in which the existing orderings (\eg horizontal, vertical or Hilbert curve \cite{bappy2019hybrid}) cannot correlate well between patches (\ie neighbor each other being separated). In contrast, self-attention mechanisms can aggregate information from the entire input image, and their global computations make them more suitable than RNNs in this domain. There are also few studies which make attempts to bring attention mechanism to image forensics \cite{hu2020span, islam2020doa}. In \cite{hu2020span}, a spatial pyramid attention network (with different dilation distances) is designed, where RGB, Bayer \cite{bayar2016deep} and SRM \cite{fridrich2012rich, zhou2017two} features are extracted. \cite{islam2020doa} using a dual-order (channel) attention module \cite{dai2019second} in GAN, which applies only to a specific manipulation type. In this work, we try to utilize self-attention to model relations between `patches' only based on RGB features for exploiting rich statistical features of different tampering artifacts. \medskip

\noindent{\bf Deep supervision}\quad Deep supervision aims at enforcing direct supervision for the hidden layers. Deeply-supervised net (DSN) \cite{lee2015deeply} makes the learning process of hidden layers transparent, which boosts the classification performance and effectively avoids the exploding and vanishing gradients. Based on this, Zhou \etal \cite{DBLP:journals/corr/abs-1807-10165} present UNet++ for medical image segmentation. These segmentation networks share a key similarity: using skip connections to combine semantic feature maps from the decoder with shallow feature maps from the encoder, which helps to improve the segmentation performance. Inspired by this, we propose a dense correction architecture for capturing both coarse-grained (high-level, semantic) and fine-grained (low-level, statistical) predictions and correcting details from different branches by multiplication (see \ref{featurefusion}). The architecture enables network pruning and producing better results.

\section{Method}

Image forensics, aiming at capturing tampered regions, is different from semantic object segmentation. For example, copy-move copies one object region to another region within the same image. In this case, the main goal of image forensics is to localize the pasted object region whereas semantic segmentation needs to segment all object regions including the original one and the manipulated one. Considering that tampering techniques can create artifacts (\eg local noise variance, boundary discrepancy), we need to design a network to capture discriminative features for finding suspicious regions in a potentially forged image.

Previous works just used hand-crafted ordering techniques to model patches relations, which cannot retain spatial information. To solve this issue, inspired by Transformer, we propose to use self-attention to learn invisible tempering artifacts hidden in tiny details of an image, which is the first attempt in image forensics. Moreover, we propose a dense correction architecture to re-correct the output, yielding excellent performance improvement.

\subsection{Self-attention for interaction modeling}

The self-attention mechanism can be used to model rich interactions between pixels or patches in an image, which provides more comprehensive and useful information for dense visual tasks. In this work, we use self-attention encoders in image forensics, which is motivated by the following observations: first, the tampering artifacts produced by different manipulation types are different, and they are commonly hidden in the details of the image; second, modeling patches relations with hand-designed patch sequence orders cannot keep spatial information of patches. If we `split' the image into $H \times W$ patches and then fed them into the self-attention encoder, all pairwise relations between patches can be extracted. This is the theoretical foundation for using self-attention encoders in tampering localization.

\begin{figure}[t]
  \centering
  \includegraphics[width=7.2cm]{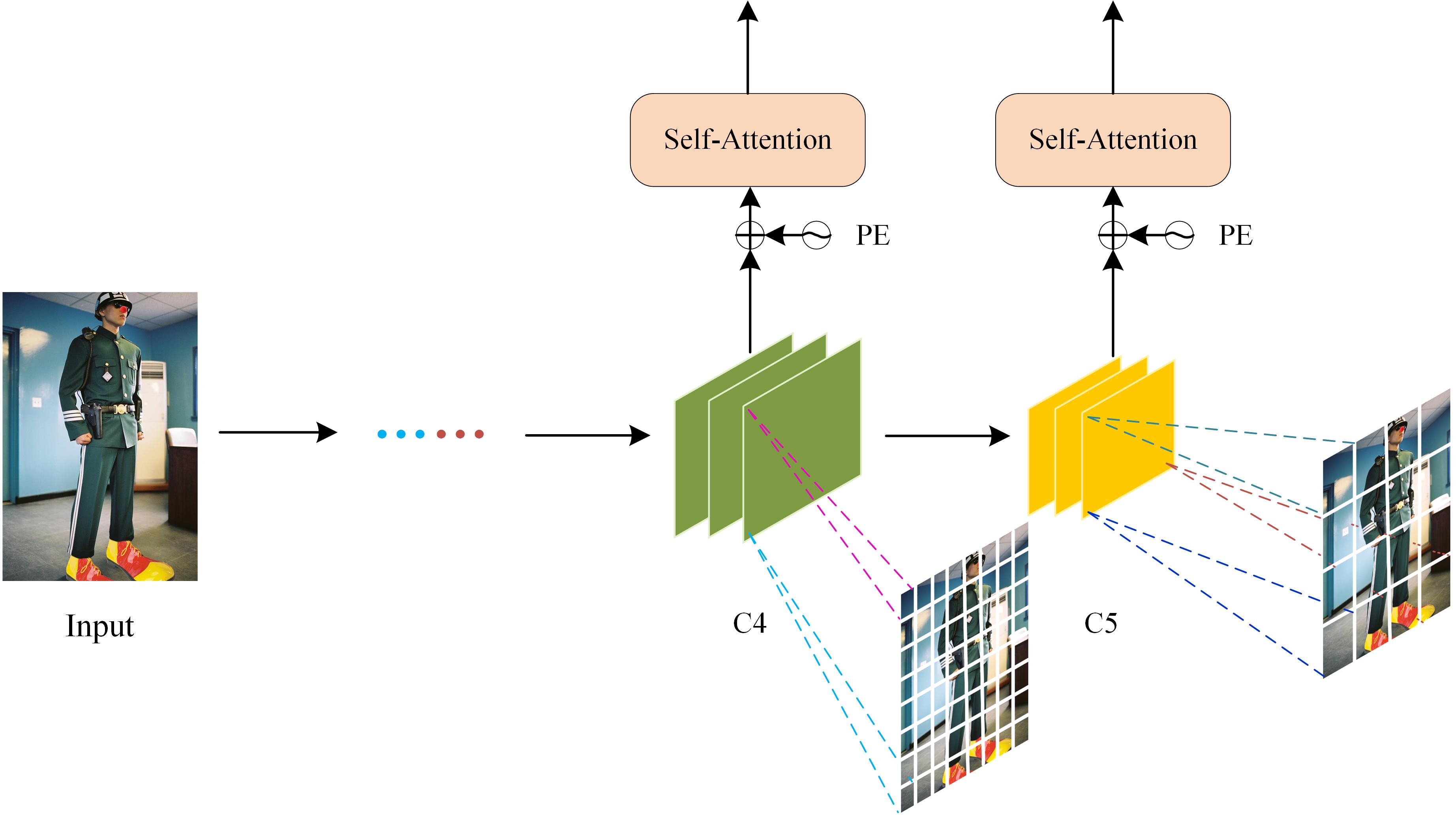}
  \caption{The correspondence between the feature map and the input image. The interactions between points in a feature map are equivalent to the relationships between patches in a digital image.}
  \label{fig:viewpatch}
\end{figure}

Each point in a feature map is equivalent to the corresponding patch in the input image, which is called respective field in deep learning (see Fig. \ref{fig:viewpatch}). So the discriminative features between patches can be extracted by modeling the relations between points in feature maps. In this paper, we do not split the whole image into a series of patches. We use ResNet50 \cite{he2016resnet} as backbone (including five stages) for feature extraction, and then we feed the outputs of the last four stages into self-attention encoders, each responsible for learning the patch relations at a different scale.

Here, we use a standard transformer encoder architecture to learn attention maps, and the details are described below. First, a $1\times1$ convolution layer is used to reduce the channel dimension of stages' output from $C$ to $d$, where $C \in [256, 512, 1024, 2048]$ and $d = 256$. Second, in order to maintain the spatial location of patches, we supplement the features with a sine positional encoding \cite{vaswani2017attention} before passing them into transformer encoders. Third, a transformer encoder has $6$ encoder layers, and each one consists of a multi-head self-attention module and a feed forward network (FFN). The dimension of feed forward is $2048$ and the dropout during training is $0.1$. In this way, we can capture the difference at the boundary location between tampered and authentic regions.

\begin{figure}[t]
  \centering
  \includegraphics[width=7.8cm]{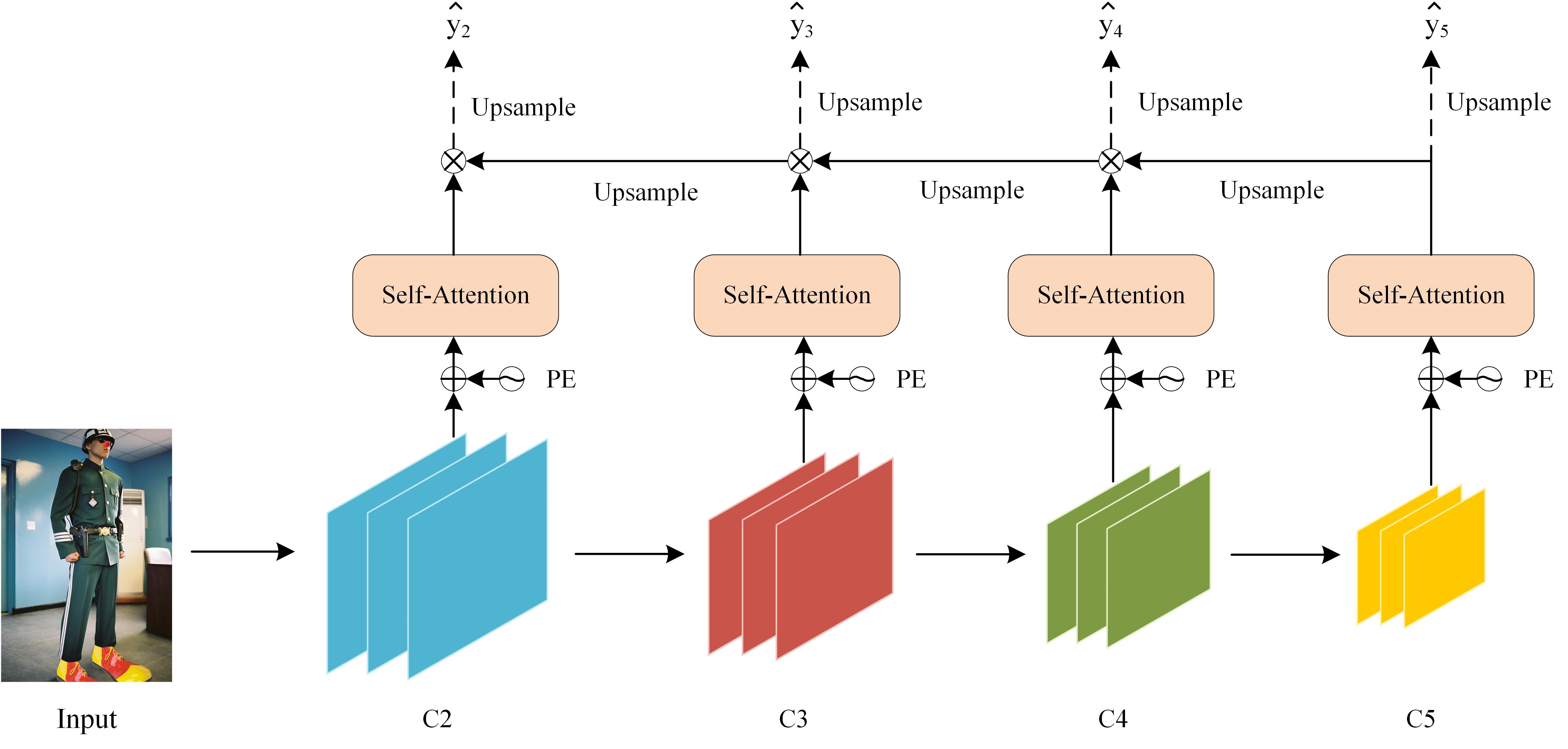}
  \caption{Network pruning. In the training phrase, comparing the results between ${\bf{\hat{y}}}_i (i = 2,...,5)$ and choosing the optimal, where the final result is produced in the test phrase.}
  \label{fig:network_pruning}
\end{figure}

\subsection{Deep supervision for network pruning}

In general, highly discriminative features will produce powerful performance. If the outputs of the hidden layers can be directly used for the final classification, the network will obtain both semantic, coarse-grained and low-level, fine-grained predictions, which contributes to the final performance. In this work, we consider adding deep supervision in the tampering localization system.

As shown in Fig. \ref{fig:network}, the network has four branches, and each output is used to calculate the localization loss separately. The advantages of this architecture are: providing more expressive features (semantic and shallow) for feature fusion (see Fig. \ref{fig:featurefusion}) and obtaining more efficient architecture by network pruning (see Fig. \ref{fig:network_pruning}). Deep supervision in this work enables the model to choose a suitable mode from all localization branches ${\bf{\hat{y}}}_i (i = 2,...,5)$, and the choice determines the extent of network pruning and speed gain (see Tab. \ref{tab:ablation3} and Tab. \ref{tab:ablation4}).

\subsection{Feature fusion for prediction correction}
\label{featurefusion}

\begin{figure}[t]
  \centering
  \includegraphics[width=5.2cm]{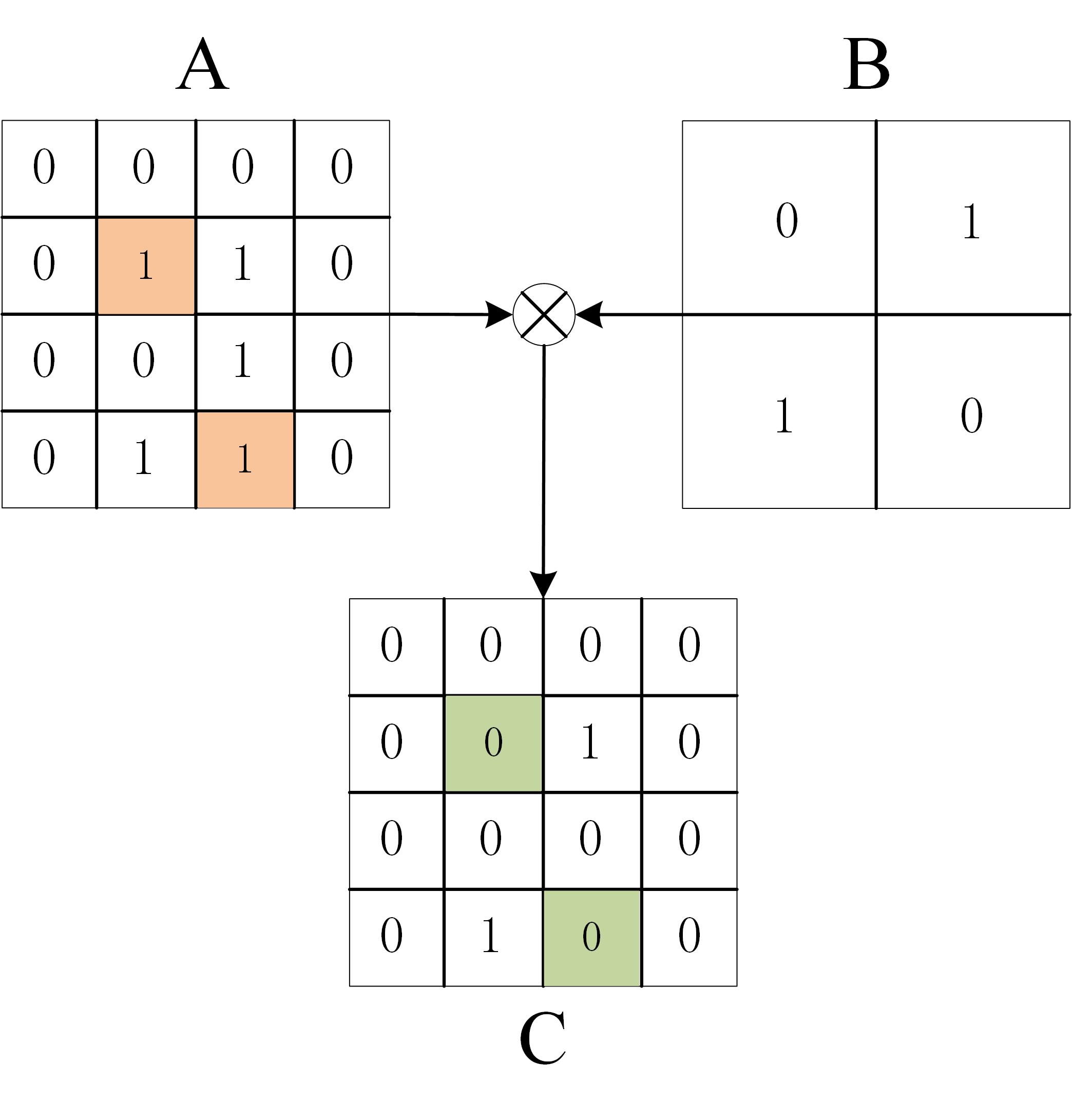}
  \caption{Feature fusion strategy. $A$ and $B$ are the outputs of two adjacent blocks, and $C$ is the result of feature fusion by multiplication. $B$ is the semantic prediction and can help to correct $A$.}
  \label{fig:featurefusion}
\end{figure}

The fully convolutional network (FCN) was introduced by Shelhamer \etal \cite{7478072} for semantic segmentation, where the upsampled features are summed with the features skipped from the encoder. Experiments show that it is effective in helping recover the full spatial resolution at the model's output. In this work, we bring a similar but new feature fusion strategy to the system. Instead of adding the features, we opt to use the multiplication operation (see Fig. \ref{fig:network} and Fig. \ref{fig:featurefusion}), and results show that it is a better choice for the tampering localization task (see Tab. \ref{tab:ablation2}).

We fuse the upsampled output from current block with the output from adjacent previous block, where they have the same size. As shown in Fig. \ref{fig:featurefusion}, $B$ represents the output of the high-level block, $A$ represents the output of the adjacent low-level block, and $C$ is the result of fusing $A$ and $B$ by multiplication. Specifically, in feature fusion modules, a $1 \times 1$ convolution is used to change the dimension of features from different branches, and the upsampling operation followed by a sigmoid function with threshold $0.5$ produces the fusing weights. The final mask prediction is computed by a $3 \times 3$ convolution with stride $1$ and padding $1$ after feature fusion by multiplication.

\subsection{Prediction loss}
\label{Prediction loss}

In this work, we use DICE loss \cite{2016V} and Focal loss \cite{lin2017focal} to supervise each mask prediction:

\begin{equation}
L_{dice}({\bf{y, \hat{y}}}) = 1 - \frac{2 \cdot \sum_k(y_k \cdot \hat{y}_k)}{\sum_k(y_k + \hat{y}_k)}
\label{con:loss1}
\end{equation}

\begin{equation}
L_{focal}(p_t) = - \alpha_t (1 - p_t) ^{\gamma} \log (p_t)
\label{con:loss2}
\end{equation}

\noindent where ${\bf{y}}$ and ${\bf{\hat{y}}}$ are ground-truth (GT) and predicted mask, and $k$ denotes the point of the mask. Let $\{\pm 1\}$ be the GT class and $p \in [0, 1]$ be the probability ($p_t = p$ for class $1$ and $p_t = 1 - p$ for class $-1$). A weighting factor $\alpha \in [0, 1]$ is introduced for addressing class imbalance ($\alpha_t = \alpha$ for class $1$ and $\alpha_t = 1 - \alpha$ for class $-1$). $\gamma$ is the tunable focusing parameter. Like \cite{lin2017focal}, we set $\alpha = 0.25$ and $\gamma = 2$. We compute the joint loss of all branches during training:

\begin{equation}
Loss = \sum_i \lambda_i (L_{dice} ({\bf{y}}_i, {\bf{\hat{y}}}_i) + L_{focal}(p_t))
\end{equation}

\noindent here, the loss for the branch $i$ is $L_{dice} ({\bf{y}}_i, {\bf{\hat{y}}}_i) + L_{focal}(p_t)$ and the corresponding ratio is $\lambda_i$. There are two ways to calculate the branch loss: upsampling predicted mask or downsampling GT mask. ${\bf{\hat{y}}}_i$ can be used to compute the branch loss directly in which the corresponding ${\bf{y}}_i$ can be obtained by downsampling the original GT mask. We can also employ upsampled mask prediction, in which ${\bf{\hat{y}}}_i$ is upsampled to have the same size as the original GT mask. Note that these two computation types are different. In upsampling, the higher the layer, the more noise may be introduced into the network via ‘nearest’ interpolating, and in this way, the loss between GT and predicted mask needs more attention, whereas downsampling does the opposite. For producing more precise predictions, we choose different coefficients for four branches to control the correction amplitude during training. Thus, the ratio set is written as follows.

\begin{equation}
\label{uratio}
{Upsampling}:\ \lambda_i < \lambda_j, i < j
\end{equation}

\begin{equation}
\label{dratio}
{Downsampling}:\ \lambda_i > \lambda_j, i < j
\end{equation}

\noindent here, we denote the low-level branch as $i$ and the high-level branch as $j$. $\lambda_i$ and $\lambda_j$ are the corresponding ratio of the branch $i$ and $j$ for calculating the joint loss.

\section{Experiments}

\subsection{Datasets}

In this work, we choose three common datasets in image forensics: CASIA \cite{dong2013casia}, COVERAGE \cite{wen2016coverage} and IMD2020 \cite{novozamsky2020imd2020}. The details are as follows. \medskip

\noindent{\bf CASIA} focuses on splicing and copy-move. It provides binary GT masks of tampered regions. The dataset includes CASIA v1.0 about 921 tampered images and CASIA v2.0 about 5123 tampered images.\medskip

\noindent{\bf COVERAGE} is a relatively small dataset designed for copy-move. It provides 100 manipulated images and corresponding GT masks.\medskip

\noindent{\bf IMD2020} is a `real-life' manipulated dataset made by unknown people and collected from the Internet. The images having obvious traces of digital manipulation are discarded and binary masks localizing tampered regions are created manually. It includes 2010 examples.

In data preparation, we split the entire dataset into three subsets and the ratio of training, validation and testing is 8:1:1. Note that these subsets are chosen randomly.

\subsection{Details}

The proposed network contains three main components: a FCN backbone for feature extraction, dense self-attention encoders for relations modeling between patches at different scales, and dense correction modules for further performance optimization. Specifically, we use ResNet-50 as backbone. We train the network with Adam setting the initial learning rate to 1e-4. We resize the input images to $512 \times 512$, and use random horizontal flipping as the only data augmentation method. To train the model, we use Pytorch $1.6.0$ to define the localization network and utilize multi-GPU setting. We set the batch size to 2 and train the model on two NVIDIA Tesla V100 GPUs over 50 epochs for different datasets. We choose the loss weights within $[0.1, 0.2, 0.3, 0.4]$ during training.

Note that some previous works chose to finetune their models trained on other bigger image forgery datasets to obtain the performance on COVERAGE, because the number of COVERAGE is small. In our experiments, we do not perform such finetuning on COVERAGE, and the results are still comparable to the SOTA methods (see the COVER column in Tab. \ref{tab:auc} and Tab. \ref{tab:F1}).\medskip

\noindent{\bf Evaluation}\quad We evaluate our model at pixel level with the benchmark metrics: $F_{1}$ score and area under curve (AUC). The higher value indicates that the performance is better.\medskip

\noindent{\bf Baseline models}\quad In this paper, we compare our work with various baseline methods. Some methods are described in \cite{zhou2018learning}, such as ELA \cite{krawetz2007picture}, NOI1 \cite{mahdian2009using}, CFA1 \cite{ferrara2012image}, J-LSTM \cite{bappy2017exploiting}, RGB-N \cite{zhou2018learning}, and other methods such as BLK \cite{li2009passive}, ADQ1 \cite{lin2009fast}, ManTra-Net \cite{wu2019mantra}, LSTM-EnDec \cite{bappy2019hybrid} and SPAN \cite{hu2020span} are described below.

\begin{itemize}[topsep=0pt,itemsep=0pt,leftmargin=*]
  \item BLK: The work focuses on extracting block artifact grids caused by the blocking processing during JPEG compression, and then detecting them with a marking procedure.
  \item ADQ1: The method aims to detect tampered images by examining the double quantization effect hidden among the DCT coefficients in JPEG images. It is insensitive to different kinds of forgery methods.
  \item ManTra-Net: A unified deep neural architecture performing both detection and localization can handle many known forgery types.
  \item LSTM-EnDec: A manipulation localization architecture utilizing resampling features, LSTM cells and encoder-decoder modules is performed to segment out manipulated regions from non-manipulated ones.
  \item SPAN: The paper presents a spatial pyramid attention network, where Bayer and SRM features are extracted.
\end{itemize}

\subsection{Results}

In this subsection, we compare results of various works on three standard datasets quantitatively and qualitatively.\medskip

\noindent{\bf Quantitative analysis}\quad We compare our work with other various SOTA models on the datasets mentioned above with the benchmark metric AUC and $F_{1}$. From Tab. \ref{tab:auc} and Tab. \ref{tab:F1}, we can see that our proposed network outperforms baseline models by a large margin. The results of our method through upsampling and downsampling are comparable in AUC metric and the former is better in $F_{1}$ score. Note that IMD2020 was released in 2020, and we struggled to find the published literatures reporting $F_{1}$ score on this dataset, but in vain until we submitted the paper. The $F_{1}$ score of our method on IMD2020 is 0.545. About cross-dataset results: training our network with IMD2020, the AUC performance on CASIA and COVER are 0.652 and 0.758, respectively. It proves the generalizability of our method.

\begin{table}[hp]
\begin{center}
\begin{tabular}{|l|ccc|}
\hline
Method	& CASIA & COVER & IMD2020 \\
\hline\hline
{ELA \cite{krawetz2007picture}} & 0.613 & 0.583 & - \\
{NOI1 \cite{mahdian2009using}} & 0.612 & 0.587 & - \\
{CFA1 \cite{ferrara2012image}} & 0.522 & 0.485 & 0.586 \\
{J-LSTM \cite{bappy2017exploiting}} & - & 0.614 & 0.487 \\
{RGB-N \cite{zhou2018learning}} & 0.795 & 0.817 & - \\
{BLK \cite{li2009passive}} & - & - & 0.596 \\
{ADQ1 \cite{lin2009fast}} & - & - & 0.579 \\
{ManTra-Net \cite{wu2019mantra}} & 0.817 & 0.819 & 0.748 \\
{LSTM-EnDec \cite{bappy2019hybrid}} & - & 0.712 & - \\
{Ours (downsample)} & {\bf 0.850} & {\bf 0.884}* & 0.847 \\
{Ours (upsample)} & 0.837 & 0.883* & {\bf 0.848} \\
\hline
\end{tabular}
\end{center}
\caption{AUC performance comparison against different works on image forgery localization. `*' denotes that our experiments on COVERAGE do not perform finetuning, which is different from the other methods in table. `-' denotes that the result is not available in the literature.}
\label{tab:auc}
\end{table}

\begin{table}[hp]
\begin{center}
\begin{tabular}{|l|cc|}
\hline
Method	& CASIA & COVER \\
\hline\hline
{ELA \cite{krawetz2007picture}} & 0.214 & 0.222 \\
{NOI1 \cite{mahdian2009using}} & 0.263 & 0.269 \\
{CFA1 \cite{ferrara2012image}} & 0.207 & 0.190 \\
{RGB-N \cite{zhou2018learning}} & 0.408 & 0.437 \\
{SPAN \cite{hu2020span}} & 0.382 & 0.558 \\
{Ours (downsample)} & 0.479 & 0.648* \\
{Ours (upsample)} & {\bf 0.627} & {\bf 0.674}* \\
\hline
\end{tabular}
\end{center}
\caption{$F_{1}$ score performance comparison against different works. `*' and `-' have the same meaning as Tab. \ref{tab:auc}.}
\label{tab:F1}
\end{table}

\begin{figure*}[t]
  \centering
  \includegraphics[width=17.2cm]{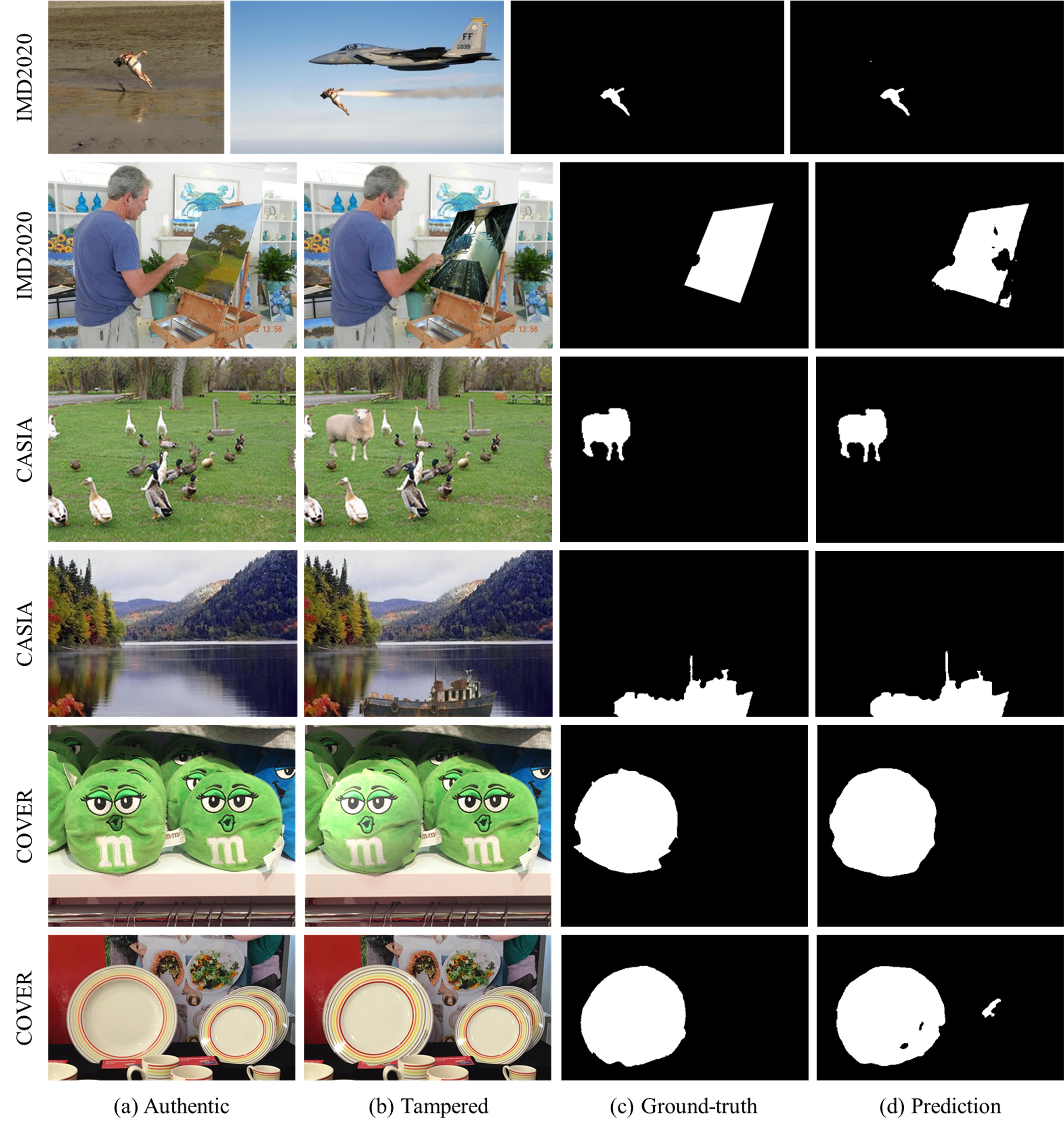}
  \caption{Examples of mask predictions using the proposed dense attention network. Images are taken from three common datasets mentioned above. From left to right: (a) authentic images, (b) tampered images, (c) ground-truth masks, and (d) mask predictions.}
  \label{fig:results}
\end{figure*}

\noindent{\bf Qualitative analysis}\quad After training, our model can generate high quality mask predictions that depict tampering locations. Here, we provide some qualitative examples (see Fig. \ref{fig:results}), which are taken from the datasets mentioned above. These examples are generated by copy-move, splicing, and `real-life' tampering from the Internet. As we can see, our method can find tampering regions with different types of manipulation. Note that the predictions are produced directly by our model without any post-processing.

\begin{table*}[htp]
\begin{center}
\begin{tabular}{|ccccccc|}
\hline
{FCN} & {Self-attention} & {Positional encoding} & {Dense correction} & CASIA & COVER & IMD2020 \\
\hline\hline
$\surd$ & $ $ & $ $ & $ $ & 0.669 & 0.715 & 0.735 \\
$\surd$ & $\surd$ & $ $ & $ $ & 0.655 & 0.674 & 0.725 \\
$\surd$ & $ $ & $\surd$ & $ $ & 0.650 & 0.707 & 0.729 \\
$\surd$ & $\surd$ & $\surd$ & $ $ & 0.809 & 0.862 & 0.815 \\
$\surd$ & $\surd$ & $\surd$ & $\surd$ & {\bf 0.837} & {\bf 0.883} & {\bf 0.848} \\
\hline
\end{tabular}
\end{center}
\caption{AUC performance in ablations.}
\label{tab:ablation1}
\end{table*}

\subsection{Ablations}

In this subsection, we conduct ablation experiments to study how the components of our proposed architecture influence the localization performance. For the ablation analysis, we use a FCN (\ie ResNet50) as backbone and evaluate the importance of the self-attention, sine spatial positional encoding, and dense correction by adding the corresponding module respectively. We also provide a detailed study to show how the choice of feature fusion and network pruning affect the final performance. \medskip

\noindent{\bf Architecture analysis}\quad The self-attention mechanism is the key component for modeling rich interactions between set elements, where the positional encoding is very important. As shown in Tab. \ref{tab:ablation1}, performance is significantly improved only when self-attention and positional encoding are used at the same time. Dense correction improves the transparency of the hidden layers, which minimizes the loss error more effectively, and re-corrects the mask prediction based on semantic dependencies between different attention maps. From the last line in Tab. \ref{tab:ablation1}, we can observe that dense correction is helpful for the final performance. \medskip

\noindent{\bf Network pruning}\quad Dense correction makes network pruning possible. As shown in see Fig. \ref{fig:network_pruning}, we can compare the results between ${\bf{\hat{y}}}_i (i = 2,...,5)$ in the training phase and choosing the optimal, where the final result is produced in the test phrase. Tab. \ref{tab:ablation3} shows the AUC performance of different branches of the proposed architecture, and Tab. \ref{tab:ablation4} provides the information of time spent and GPU memory occupancy. As can be seen in Tab. \ref{tab:ablation3}, the low-level features contribute more to the final localization performance than deep ones, which means that our model can learn invisible traces that are often hidden in tiny details of the images. Combining Tab. \ref{tab:ablation3} and Tab. \ref{tab:ablation4}, we can see that $C_3$ branch (\ie ${\bf{\hat{y}}}_3$) outperforms the other branches both in performance and time-consuming relatively.

\begin{table}[hp]
\begin{center}
\begin{tabular}{|c|ccc|}
\hline
Method	& CASIA & COVER & IMD2020 \\
\hline\hline
{Ours ($C_2$)} & 0.837 & 0.883 & 0.848 \\
{Ours ($C_3$)} & 0.835 & 0.887 & 0.847 \\
{Ours ($C_4$)} & 0.835 & 0.880 & 0.837 \\
{Ours ($C_5$)} & 0.826 & 0.877 & 0.827 \\
\hline
\end{tabular}
\end{center}
\caption{AUC performance comparison using different branches as final output (see Fig. \ref{fig:network_pruning}).}
\label{tab:ablation3}
\end{table}

\begin{table}[h]
\begin{center}
\begin{tabular}{|c|cc|}
\hline
Method	& {GPU memory (M)} & {Inference time (s)} \\
\hline\hline
{Ours ($C_2$)} & 6687 & 0.11 \\
{Ours ($C_3$)} & 774 & 0.09 \\
{Ours ($C_4$)} & 368 & 0.08 \\
{Ours ($C_5$)} & 353 & 0.07 \\
\hline
\end{tabular}
\end{center}
\caption{Comparisons of results of different branches as final output in GPU memory and inference time.}
\label{tab:ablation4}
\end{table}

During training, with the help of dense correction, the features with low-spatial resolution can get feedback including edges from low-level high-resolution features. This is the most significant difference from networks with single direction connection. Thus, although the prediction of $C_3$ branch is block-wise, it does not lack refinement on edges. \medskip

\noindent{\bf Type of feature fusion}\quad As described above, feature fusion has two types: multiplication and addition. In our experiments, we try both addition and multiplication (see Tab. \ref{tab:ablation2}), and experimental results demonstrate that the multiplication type is more suitable for the tampering localization task, which is consistent with expectations.

\begin{table}[hp]
\begin{center}
\begin{tabular}{|c|ccc|}
\hline
Method	& CASIA & COVER & IMD2020 \\
\hline\hline
{Ours (Mul)} & {\bf 0.837} & {\bf 0.883} & {\bf 0.848} \\
{Ours (Add)} & 0.739 & 0.857 & 0.827 \\
\hline
\end{tabular}
\end{center}
\caption{AUC comparison using different types for feature fusion.}
\label{tab:ablation2}
\end{table}

\section{Conclusion}

We present TransForensics, which uses dense self-attention encoders to model global context and all pairwise interactions between patches at different scales. It is the first work that introduces self-attention mechanisms of transformers to localizing tampered regions. Further, dense correction modules re-correct mask predictions by multiplication for nicer results. We demonstrate the system's ability to detect tampering artifacts for diverse realistic tampered images and to achieve a balance between performance and time-consuming. Experiments show that the proposed system can provide a powerful model for image forgery. Our method can be served as a solid but simple-to-implement baseline for image forensics, and can also be employed as a defense against data poisoning attacks to protect our learning system. In future work, the dense self-attention architecture would be a novel approach in other tasks, such as object detection and semantic segmentation.

\clearpage
{\small
\bibliographystyle{ieee_fullname}
\bibliography{reference}

\begin{thebibliography}{10}\itemsep=-1pt

\bibitem{amerini2017localization}
Irene Amerini, Tiberio Uricchio, Lamberto Ballan, and Roberto Caldelli.
\newblock Localization of jpeg double compression through multi-domain
  convolutional neural networks.
\newblock In {\em 2017 IEEE Conference on computer vision and pattern
  recognition workshops (CVPRW)}, pages 1865--1871. IEEE, 2017.

\bibitem{bammey2020adaptive}
Quentin Bammey, Rafael Grompone~von Gioi, and Jean-Michel Morel.
\newblock An adaptive neural network for unsupervised mosaic consistency
  analysis in image forensics.
\newblock In {\em Proceedings of the IEEE/CVF Conference on Computer Vision and
  Pattern Recognition}, pages 14194--14204, 2020.

\bibitem{bappy2017exploiting}
Jawadul~H Bappy, Amit~K Roy-Chowdhury, Jason Bunk, Lakshmanan Nataraj, and BS
  Manjunath.
\newblock Exploiting spatial structure for localizing manipulated image
  regions.
\newblock In {\em Proceedings of the IEEE international conference on computer
  vision}, pages 4970--4979, 2017.

\bibitem{bappy2019hybrid}
Jawadul~H Bappy, Cody Simons, Lakshmanan Nataraj, BS Manjunath, and Amit~K
  Roy-Chowdhury.
\newblock Hybrid lstm and encoder--decoder architecture for detection of image
  forgeries.
\newblock {\em IEEE Transactions on Image Processing}, 28(7):3286--3300, 2019.

\bibitem{barni2017aligned}
Mauro Barni, Luca Bondi, Nicol{\`o} Bonettini, Paolo Bestagini, Andrea
  Costanzo, Marco Maggini, Benedetta Tondi, and Stefano Tubaro.
\newblock Aligned and non-aligned double jpeg detection using convolutional
  neural networks.
\newblock {\em Journal of Visual Communication and Image Representation},
  49:153--163, 2017.

\bibitem{bayar2016deep}
Belhassen Bayar and Matthew~C Stamm.
\newblock A deep learning approach to universal image manipulation detection
  using a new convolutional layer.
\newblock In {\em Proceedings of the 4th ACM workshop on information hiding and
  multimedia security}, pages 5--10, 2016.

\bibitem{bianchi2011improved}
Tiziano Bianchi, Alessia De~Rosa, and Alessandro Piva.
\newblock Improved dct coefficient analysis for forgery localization in jpeg
  images.
\newblock In {\em 2011 IEEE International Conference on Acoustics, Speech and
  Signal Processing (ICASSP)}, pages 2444--2447. IEEE, 2011.

\bibitem{nicolas2020transformers}
Nicolas Carion, Francisco Massa, Gabriel Synnaeve, Nicolas Usunier, Alexander
  Kirillov, and Sergey Zagoruyko.
\newblock End-to-end object detection with transformers.
\newblock In {\em 2020 IEEE European Conference on Computer Vision (ECCV)}.
  IEEE, 2020.

\bibitem{cozzolino2014image}
Davide Cozzolino, Diego Gragnaniello, and Luisa Verdoliva.
\newblock Image forgery localization through the fusion of camera-based,
  feature-based and pixel-based techniques.
\newblock In {\em 2014 IEEE International Conference on Image Processing
  (ICIP)}, pages 5302--5306. IEEE, 2014.

\bibitem{cozzolino2018noiseprint}
Davide Cozzolino and Luisa Verdoliva.
\newblock Noiseprint: a cnn-based camera model fingerprint.
\newblock {\em arXiv preprint arXiv:1808.08396}, 2018.

\bibitem{dai2019second}
Tao Dai, Jianrui Cai, Yongbing Zhang, Shu-Tao Xia, and Lei Zhang.
\newblock Second-order attention network for single image super-resolution.
\newblock In {\em Proceedings of the IEEE/CVF Conference on Computer Vision and
  Pattern Recognition}, pages 11065--11074, 2019.

\bibitem{de2013exposing}
Tiago~Jos{\'e} De~Carvalho, Christian Riess, Elli Angelopoulou, Helio Pedrini,
  and Anderson de Rezende~Rocha.
\newblock Exposing digital image forgeries by illumination color
  classification.
\newblock {\em IEEE Transactions on Information Forensics and Security},
  8(7):1182--1194, 2013.

\bibitem{dong2013casia}
Jing Dong, Wei Wang, and Tieniu Tan.
\newblock Casia image tampering detection evaluation database.
\newblock In {\em 2013 IEEE China Summit and International Conference on Signal
  and Information Processing}, pages 422--426. IEEE, 2013.

\bibitem{ferrara2012image}
Pasquale Ferrara, Tiziano Bianchi, Alessia De~Rosa, and Alessandro Piva.
\newblock Image forgery localization via fine-grained analysis of cfa
  artifacts.
\newblock {\em IEEE Transactions on Information Forensics and Security},
  7(5):1566--1577, 2012.

\bibitem{fridrich2012rich}
Jessica Fridrich and Jan Kodovsky.
\newblock Rich models for steganalysis of digital images.
\newblock {\em IEEE Transactions on Information Forensics and Security},
  7(3):868--882, 2012.

\bibitem{he2016resnet}
Kaiming He, Xiangyu Zhang, Shaoqing Ren, and Jian Sun.
\newblock Deep residual learning for image recognition.
\newblock In {\em 2016 IEEE Conference on Computer Vision and Pattern
  Recognition (CVPR)}, pages 770--778, 2016.

\bibitem{hu2020span}
Xuefeng Hu and Zhihan Zhang.
\newblock Span: Spatial pyramid attention network for image manipulation
  localization.
\newblock In {\em ECCV}, 2020.

\bibitem{islam2020doa}
Ashraful Islam, Chengjiang Long, Arslan Basharat, and Anthony Hoogs.
\newblock Doa-gan: Dual-order attentive generative adversarial network for
  image copy-move forgery detection and localization.
\newblock In {\em Proceedings of the IEEE/CVF Conference on Computer Vision and
  Pattern Recognition}, pages 4676--4685, 2020.

\bibitem{khan2021transformers}
Salman Khan, Muzammal Naseer, Munawar Hayat, Syed~Waqas Zamir, Fahad~Shahbaz
  Khan, and Mubarak Shah.
\newblock Transformers in vision: A survey.
\newblock {\em arXiv preprint arXiv:2101.01169}, 2021.

\bibitem{krawetz2007picture}
Neal Krawetz and Hacker~Factor Solutions.
\newblock A picture’s worth.
\newblock {\em Hacker Factor Solutions}, 6(2):2, 2007.

\bibitem{lee2015deeply}
Chen-Yu Lee, Saining Xie, Patrick Gallagher, Zhengyou Zhang, and Zhuowen Tu.
\newblock Deeply-supervised nets.
\newblock In {\em Artificial intelligence and statistics}, pages 562--570.
  PMLR, 2015.

\bibitem{li2020face}
Lingzhi Li, Jianmin Bao, Ting Zhang, Hao Yang, Dong Chen, Fang Wen, and Baining
  Guo.
\newblock Face x-ray for more general face forgery detection.
\newblock In {\em Proceedings of the IEEE/CVF Conference on Computer Vision and
  Pattern Recognition}, pages 5001--5010, 2020.

\bibitem{li2009passive}
Weihai Li, Yuan Yuan, and Nenghai Yu.
\newblock Passive detection of doctored jpeg image via block artifact grid
  extraction.
\newblock {\em Signal Processing}, 89(9):1821--1829, 2009.

\bibitem{lin2017focal}
Tsung-Yi Lin, Priya Goyal, Ross Girshick, Kaiming He, and Piotr Doll{\'a}r.
\newblock Focal loss for dense object detection.
\newblock In {\em Proceedings of the IEEE international conference on computer
  vision}, pages 2980--2988, 2017.

\bibitem{lin2009fast}
Zhouchen Lin, Junfeng He, Xiaoou Tang, and Chi-Keung Tang.
\newblock Fast, automatic and fine-grained tampered jpeg image detection via
  dct coefficient analysis.
\newblock {\em Pattern Recognition}, 42(11):2492--2501, 2009.

\bibitem{lyu2014exposing}
Siwei Lyu, Xunyu Pan, and Xing Zhang.
\newblock Exposing region splicing forgeries with blind local noise estimation.
\newblock {\em International journal of computer vision}, 110(2):202--221,
  2014.

\bibitem{mahdian2009using}
Babak Mahdian and Stanislav Saic.
\newblock Using noise inconsistencies for blind image forensics.
\newblock {\em Image and Vision Computing}, 27(10):1497--1503, 2009.

\bibitem{mazumdar2018universal}
Aniruddha Mazumdar, Jaya Singh, Yosha~Singh Tomar, and Prabin~Kumar Bora.
\newblock Universal image manipulation detection using deep siamese
  convolutional neural network.
\newblock {\em arXiv preprint arXiv:1808.06323}, 2018.

\bibitem{2016V}
Fausto Milletari, Nassir Navab, and Seyed~Ahmad Ahmadi.
\newblock V-net: Fully convolutional neural networks for volumetric medical
  image segmentation.
\newblock In {\em 2016 Fourth International Conference on 3D Vision (3DV)},
  2016.

\bibitem{novozamsky2020imd2020}
Adam Novozamsky, Babak Mahdian, and Stanislav Saic.
\newblock Imd2020: A large-scale annotated dataset tailored for detecting
  manipulated images.
\newblock In {\em Proceedings of the IEEE/CVF Winter Conference on Applications
  of Computer Vision Workshops}, pages 71--80, 2020.

\bibitem{pan2012exposing}
Xunyu Pan, Xing Zhang, and Siwei Lyu.
\newblock Exposing image splicing with inconsistent local noise variances.
\newblock In {\em 2012 IEEE International Conference on Computational
  Photography (ICCP)}, pages 1--10. IEEE, 2012.

\bibitem{riess2010scene}
Christian Riess and Elli Angelopoulou.
\newblock Scene illumination as an indicator of image manipulation.
\newblock In {\em International Workshop on Information Hiding}, pages 66--80.
  Springer, 2010.

\bibitem{7478072}
Evan Shelhamer, Jonathan Long, and Trevor Darrell.
\newblock Fully convolutional networks for semantic segmentation.
\newblock {\em IEEE Transactions on Pattern Analysis and Machine Intelligence},
  39(4):640--651, April 2017.

\bibitem{vaswani2017attention}
Ashish Vaswani, Noam Shazeer, Niki Parmar, Jakob Uszkoreit, Llion Jones,
  Aidan~N Gomez, {\L}ukasz Kaiser, and Illia Polosukhin.
\newblock Attention is all you need.
\newblock In {\em Advances in neural information processing systems}, pages
  5998--6008, 2017.

\bibitem{verde2020focal}
Sebastiano Verde, Paolo Bestagini, Simone Milani, Giancarlo Calvagno, and
  Stefano Tubaro.
\newblock Focal: A forgery localization framework based on video coding
  self-consistency.
\newblock {\em arXiv preprint arXiv:2008.10454}, 2020.

\bibitem{wang2016double}
Qing Wang and Rong Zhang.
\newblock Double jpeg compression forensics based on a convolutional neural
  network.
\newblock {\em EURASIP Journal on Information Security}, 2016(1):1--12, 2016.

\bibitem{wen2016coverage}
Bihan Wen, Ye Zhu, Ramanathan Subramanian, Tian-Tsong Ng, Xuanjing Shen, and
  Stefan Winkler.
\newblock Coverage—a novel database for copy-move forgery detection.
\newblock In {\em 2016 IEEE international conference on image processing
  (ICIP)}, pages 161--165. IEEE, 2016.

\bibitem{wu2019mantra}
Yue Wu, Wael AbdAlmageed, and Premkumar Natarajan.
\newblock Mantra-net: Manipulation tracing network for detection and
  localization of image forgeries with anomalous features.
\newblock In {\em Proceedings of the IEEE/CVF Conference on Computer Vision and
  Pattern Recognition}, pages 9543--9552, 2019.

\bibitem{ye2019crossmodal}
Linwei Ye, Mrigank Rochan, Zhi Liu, and Yang Wang.
\newblock Cross-modal self-attention network for referring image segmentation.
\newblock In {\em 2019 IEEE Conference on Computer Vision and Pattern
  Recognition (CVPR)}, page 10502–10511, 2019.

\bibitem{zheng2020rethinking}
Sixiao Zheng, Jiachen Lu, Hengshuang Zhao, Xiatian Zhu, Zekun Luo, Yabiao Wang,
  Yanwei Fu, Jianfeng Feng, Tao Xiang, Philip~HS Torr, et~al.
\newblock Rethinking semantic segmentation from a sequence-to-sequence
  perspective with transformers.
\newblock {\em arXiv preprint arXiv:2012.15840}, 2020.

\bibitem{zhou2017two}
Peng Zhou, Xintong Han, Vlad~I Morariu, and Larry~S Davis.
\newblock Two-stream neural networks for tampered face detection.
\newblock In {\em 2017 IEEE Conference on Computer Vision and Pattern
  Recognition Workshops (CVPRW)}, pages 1831--1839. IEEE, 2017.

\bibitem{zhou2018learning}
Peng Zhou, Xintong Han, Vlad~I Morariu, and Larry~S Davis.
\newblock Learning rich features for image manipulation detection.
\newblock In {\em Proceedings of the IEEE Conference on Computer Vision and
  Pattern Recognition}, pages 1053--1061, 2018.

\bibitem{DBLP:journals/corr/abs-1807-10165}
Zongwei Zhou, Md~Mahfuzur~Rahman Siddiquee, Nima Tajbakhsh, and Jianming Liang.
\newblock Unet++: {A} nested u-net architecture for medical image segmentation.
\newblock {\em CoRR}, abs/1807.10165, 2018.

\end{thebibliography}
}

\end{document}